\documentclass{article}

\usepackage{microtype}
\usepackage{graphicx}
\usepackage{subfigure}
\usepackage{booktabs} 

\usepackage{hyperref}

\usepackage[accepted]{icml2018}

\usepackage{amsmath}
\usepackage{graphicx}
\usepackage{dsfont}
\usepackage{booktabs}
\usepackage{multirow}
\usepackage{wrapfig}
\usepackage{float}

\icmltitlerunning{Visualizing and Understanding Atari Agents}

\begin{document}

\twocolumn[
\icmltitle{Visualizing and Understanding Atari Agents}




\begin{icmlauthorlist}
\icmlauthor{Sam Greydanus}{osu}
\icmlauthor{Anurag Koul}{osu}
\icmlauthor{Jonathan Dodge}{osu}
\icmlauthor{Alan Fern}{osu}
\end{icmlauthorlist}

\icmlaffiliation{osu}{Oregon State University, Corvallis, Oregon, USA}

\icmlcorrespondingauthor{Sam Greydanus}{greydanus.17@gmail.com}

\icmlkeywords{reinforcement learning, interpretability, Atari}

\vskip 0.3in
]



\printAffiliationsAndNotice{}  


\begin{abstract}
While deep reinforcement learning (deep RL) agents are effective at maximizing rewards, it is often unclear what strategies they use to do so. In this paper, we take a step toward explaining deep RL agents through a case study using Atari 2600 environments. In particular, we focus on using saliency maps to understand how an agent learns and executes a policy. We introduce a method for generating useful saliency maps and use it to show 1) what strong agents attend to, 2) whether agents are making decisions for the right or wrong reasons, and 3) how agents evolve during learning. We also test our method on non-expert human subjects and find that it improves their ability to reason about these agents. Overall, our results show that saliency information can provide significant insight into an RL agent's decisions and learning behavior.
\end{abstract}

\section{Introduction}

Deep learning algorithms have achieved state-of-the-art results in image classification \cite{He, Krizhevsky2012ImageNetNetworks}, machine translation \cite{Mikolov2010RecurrentModel}, image captioning \cite{Karpathy2015DeepDescriptions}, drug discovery \cite{Dahl2014Multi-taskPredictions}, and deep reinforcement learning \cite{Mnih2015Human-levelLearning, Silver2017MasteringKnowledge}. In spite of their impressive performance on such tasks, they are often criticized for being black boxes. Researchers must learn to interpret these models before using them to solve real-world problems where trust and reliability are critical.

While an abundance of literature has addressed how to explain deep image classifiers \cite{Fong2017InterpretablePerturbation, Ribeiro2016WhyClassifier, Simonyan2014DeepMaps, Zhang2016Top-downBackprop} and deep sequential models \cite{Karpathy2016VisualizingNetworks, Murdoch2017AutomaticNetworks}, very little work has focused on explaining deep RL agents. These agents are known to perform well in challenging environments that have sparse rewards and noisy, high-dimensional inputs. Simply observing the policies of these agents is one way to understand them. However, explaining their decision-making process in more detail requires better tools.

In this paper, we investigate deep RL agents that use raw visual input to make their decisions. In particular, we focus on exploring the utility of visual saliency to gain insight into the decisions made by these agents. To the best of our knowledge, there has not been a thorough investigation of saliency for this purpose. Thus, it is unclear which saliency methods produce meaningful visualizations across full episodes of an RL agent and whether those visualizations yield insight.

Past methods for visualizing deep RL agents include t-SNE embeddings \cite{Mnih2015Human-levelLearning, Zahavy2016GrayingDQNs}, Jacobian saliency maps \cite{Wang2016DuelingLearning, Zahavy2016GrayingDQNs}, and reward curves \cite{Mnih2015Human-levelLearning}. These tools are difficult for non-experts to interpret, due to the need to understand expert-level concepts, such as embeddings. Other tools, such as reward curves, treat the agents as black boxes and hence provide limited explanatory power about the internal decision making processes. Our  work is motivated by trying to strike a favorable balance between interpretability and insight into the underlying decision making.

\begin{figure}[h!]
\centering
\includegraphics[width=0.48\textwidth]{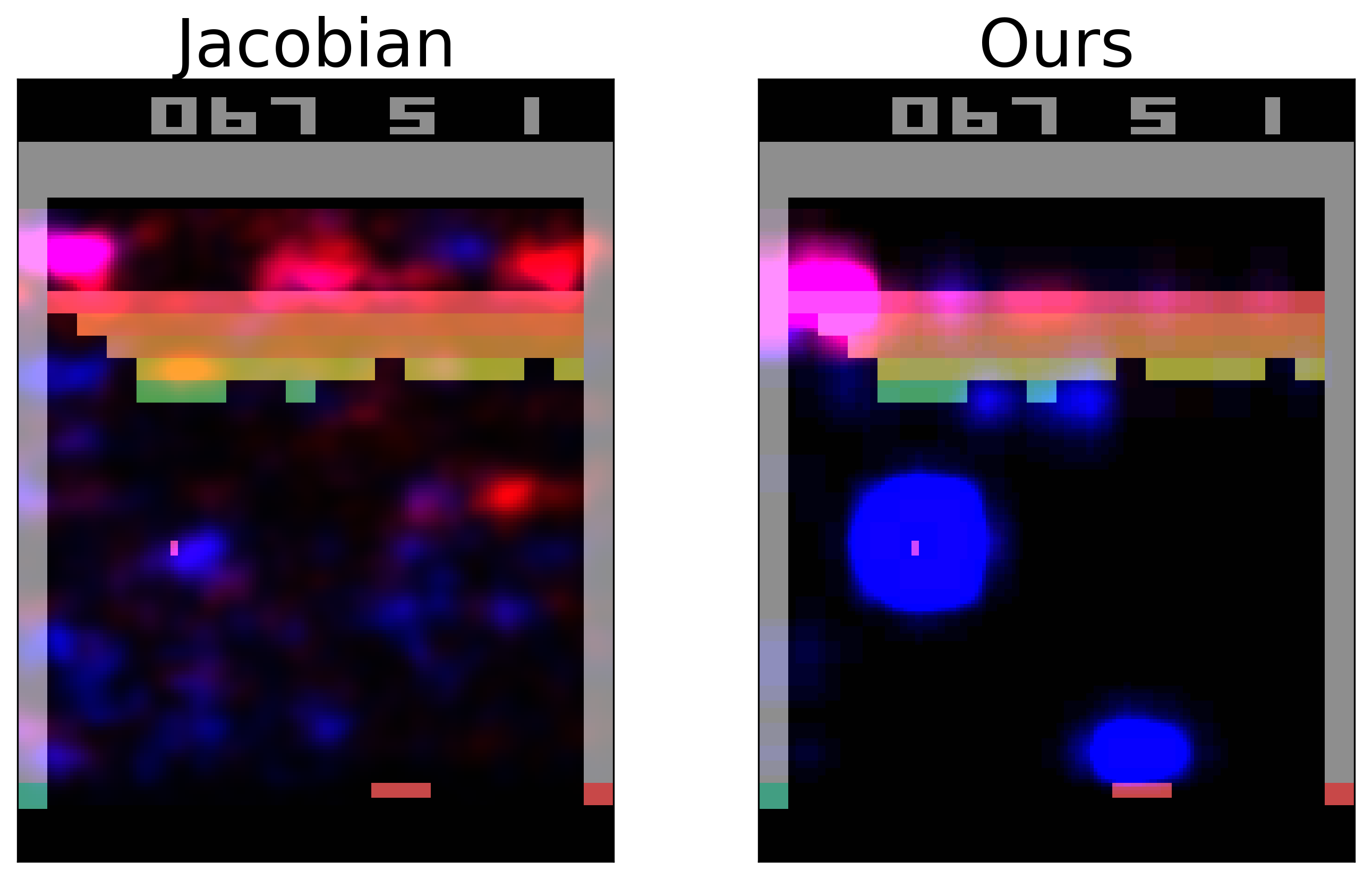}
\caption{Comparison of Jacobian saliency to our perturbation-based approach. We are visualizing an actor-critic model \cite{Mnih2016AsynchronousLearning}. Red indicates saliency for the critic; blue is saliency for the actor.}
\label{fig:jac-vs-pert}
\end{figure}

Our first contribution is to describe a simple perturbation-based technique for generating saliency videos of deep RL agents. Our work was motivated by the generally poor quality of Jacobian saliency, which has been the primary visualization tool for deep RL agents in prior work (see Figure \ref{fig:jac-vs-pert}). For the sake of thoroughness, we limit our experiments to six Atari 2600 environments: Pong, SpaceInvaders, Breakout, MsPacman, Frostbite, and Enduro. Our long-term goal is to visualize and understand the policies of \textit{any} deep reinforcement learning agent that uses visual inputs. We make our code and results available online\footnote{\texttt{github.com/greydanus/visualize\_atari}}.

Our main contribution is to conduct a series of investigative explorations into explaining Atari agents. First, we identify the key strategies of the three agents that exceed human baselines in their environments. Second, we visualize agents throughout training to see how their policies evolved. Third, we explore the use of saliency for detecting when an agent is earning high rewards for the ``wrong reasons". This includes a demonstration that the saliency approach allows non-experts to detect such situations. Fourth, we consider Atari games where trained agents perform poorly. We use saliency to ``debug'' these agents by identifying the basis of their low-quality decisions. 

Most of our paper focuses on understanding how an agent's current state affects its current policy. However, since we use an agent with recurrent structure, we acknowledge that memory is also important. A simple example is an agent which has learned to reason about the velocity of a ball; it uses information about previous frames \textit{in addition to} information from the current frame. In response to these concerns, we present preliminary experiments on visualizing the role of memory.

\section{Related Work}

\textbf{Explaining traditional RL agents.} Prior work has generated natural language and logic-based explanations for policies in Markov Decision Processes (MDP) \cite{Dodson2011AProcesses, Elizalde2008PolicyProcesses, Khan2009MinimalProcesses}. These methods assume access to an exact MDP model (e.g. represented as a dynamic Bayesian network) and that the policies map from interpretable, high-level state features to actions. Neither assumption is valid in our vision-based domain.

More recently, there has been work on analyzing execution traces of an RL agent in order to extract explanations \cite{Hayes2017ImprovingExplanation}. A problem with this approach is that it relies heavily on hand-crafted state features which are semantically meaningful to humans. This is impractical for vision-based applications, where agents must learn directly from pixels.

\textbf{Explaining deep RL agents.} Recent work by Zahavy et al. \cite{Zahavy2016GrayingDQNs} has developed tools for explaining deep RL policies in visual domains. Similar to our work, the authors use the Atari 2600 environments as interpretable testbeds. Their key contribution is a method of approximating the behavior of deep RL policies via Semi-Aggregated Markov Decision Processes (SAMDPs). They use the more interpretable SAMDPs to gain insights about the higher-level temporal structure of the policy.

While this process produces valuable insights, the analysis operates externally to the deep policy and hence does not provide insights into the perceptual aspects of the policy. From a user perspective, an issue with the explanations is that they emphasize t-SNE clusters and state-action statistics which are uninformative to those without a machine learning background. To build user trust, it is important that explanations be obtained directly from the original policy and that they be interpretable to the untrained eye. 

Whereas work by Zahavy et al. (2016) takes a black box approach (using SAMDPs to analyze high-level policy behavior), we aim to obtain visualizations of how inputs influence individual decisions. To do this, we turned to previous literature on visual explanations of Deep Neural Networks (DNNs). We found that the most interpretable explanations generally took the form of saliency maps. While techniques varied from work to work, most fell into two main categories: gradient-based methods and perturbation-based methods.

\textbf{Gradient-based saliency methods.} Gradient methods aim to understand what features of a DNN's input are most salient to its output by using variants of the chain rule. The simplest approach is to take the Jacobian with respect to the output of interest \cite{Simonyan2014DeepMaps}. Unfortunately, the Jacobian does not usually produce human-interpretable saliency maps. Thus several variants have emerged, aimed at modifying gradients to obtain more meaningful saliency. These variants include Guided Backpropagation \cite{Springenberg2015StrivingNet}, Excitation Backpropagation \cite{Zhang2016Top-downBackprop}, and DeepLIFT \cite{Shrikumar2017LearningDifferences}.

Gradient methods are efficient to compute and have clear semantics (for input $x$ and function $f(x)$, $\frac{\partial f(x)}{\partial x_i}$ is a mathematical definition of saliency), but their saliency maps can be difficult to interpret. This is because, when answering the question \textit{``What perturbation to the input increases a particular output?"}, gradient methods can choose perturbations which lack physical meaning. Changing an input in the direction of the gradient tends to move it off from the manifold of realistic input images.

\begin{figure*}[h!]
\centering
\includegraphics[width=1.6\columnwidth]{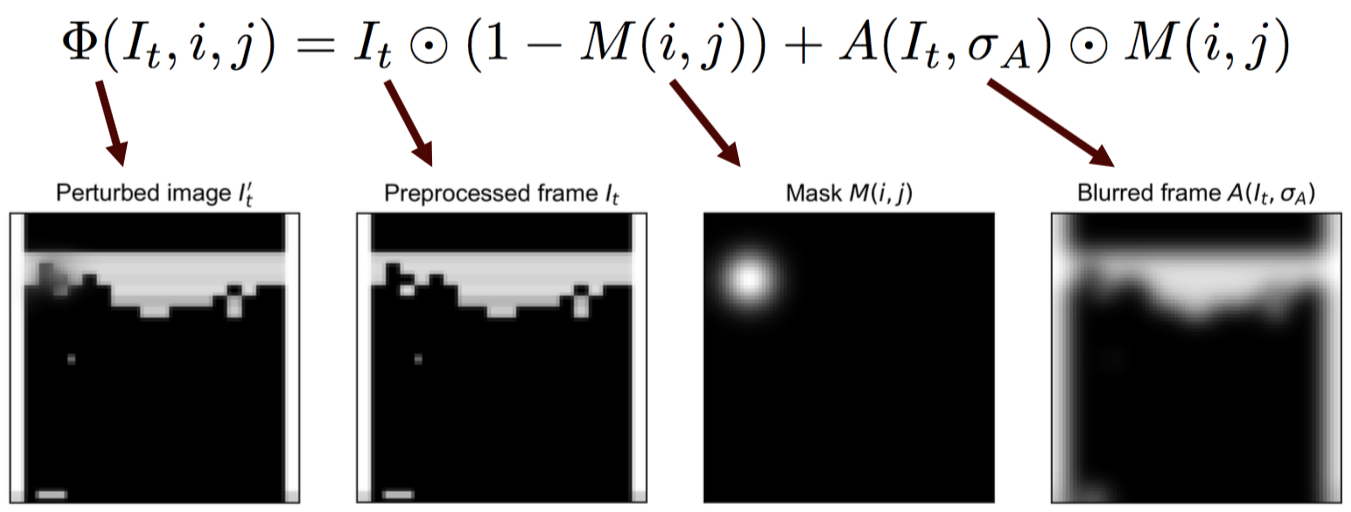}
\caption{An example of how our perturbation method selectively blurs a region, applied to a cropped frame of Breakout}
\label{fig:blur}
\end{figure*}

\textbf{Perturbation-based saliency methods.} The idea behind perturbation-based methods is to measure how a model's output changes when some of the input information is altered. For a simple example, borrowed from \cite{Fong2017InterpretablePerturbation}, consider a classifier which predicts +1 if the image contains a robin and -1 otherwise. Removing information from the part of the image which contains the robin should change the model's output, whereas doing so for other areas should not. However, choosing a perturbation which removes information without introducing any \textit{new} information can be difficult.

The simplest perturbation is to replace part of an input image with a gray square \cite{Zeiler2014VisualizingNetworks} or region \cite{Ribeiro2016WhyClassifier}. A problem with this approach is that replacing pixels with a constant color introduces unwanted color and edge information. For example, adding a gray square might increase a classifier's confidence that the image contains an elephant. More recent approaches by \cite{Dabkowski2017RealClassifiers} and \cite{Fong2017InterpretablePerturbation} use masked interpolations between the original image $I$ and some other image $A$, where $A$ is chosen to introduce as little new information as possible.

\section{Visualizing Saliency for Atari Agents}

In this work, we focus on agents trained via the Asynchronous Advantage Actor-Critic (A3C) algorithm, which is known for its ease of use and strong performance in Atari environments \cite{Mnih2016AsynchronousLearning}. A3C trains agents that have both a policy (actor) distribution $\pi$ and a value (critic) estimate $V^{\pi}$. In particular, letting $I_{1:t}$ denote the sequence of image frames from time 1 to time $t$, $\pi(I_{1:t})$ returns a distribution over actions to take at time $t$ and $V^{\pi}(I_{1:t})$ estimates the expected future value of following $\pi$ after observing $I_{1:t}$. We use a single DNN architecture to estimate both $\pi$ and $V^{\pi}$ as detailed in Section \ref{sec:experiments}.

We are interested in understanding these deep RL agents in terms of the information they use to make decisions and the relative importance of visual features. To do this, we found it useful to construct and visualize saliency maps for both $\pi$ and $V^{\pi}$ at each time step. In particular, the saliency map for $\pi(I_{1:t})$ is intended to identify the key information in frame $I_t$ that the policy uses to select action $a_t$. Similarly, the saliency map for $V^{\pi}(I_{1:t})$ is intended to identify the key information in frame $I_t$ for assigning a value at time $t$. 



\textbf{Perturbation-based saliency.} Here we introduce a perturbation which produces rich, insightful saliency maps\footnote{Saliency \textit{videos} at \texttt{https://goo.gl/jxvAKn}.}. Given an image $I_t$ at time $t$, we let $\Phi(I_t,i,j)$ denote the perturbation $I_t$ centered at pixel coordinates $(i,j)$. We define $\Phi(I_t,i,j)$ in Equation \ref{eqn:masked-interp-ours}; it is a blur localized around $(i,j)$. We construct this blur using the Hadamard product, $\odot$, to interpolate between the original image $I_t$ and a Gaussian blur, $A(I_t, \sigma_A=3)$, of that image. The interpolation coefficients are given by image mask $M(i,j) \in (0,1)^{m\times n}$ which corresponds to a 2D Gaussian centered at $\mu=(i,j)$ with $\sigma^2=25$. Figure \ref{fig:blur} shows an example of this perturbation.

\begin{equation}
\Phi(I_t,i,j) = I_t \odot (1-M(i,j)) + A(I_t, \sigma_A) \odot M(i,j)
\label{eqn:masked-interp-ours}
\end{equation}

We interpret this perturbation as adding spatial uncertainty to the region around $(i,j)$. For example, if location $(i,j)$ coincides with the location of the ball in the game of Pong, our perturbation diffuses the ball's pixels, making the policy less certain about the ball's location. 

We are interested in answering the question, \textit{``How much does removing information from the region around location $(i,j)$ change the policy?"} Let $\pi_u(I_{1:t})$ denote the logistic units, or logits, that are the inputs to the final softmax activation\footnote{Logits, in lieu of softmax output $\pi$ gave sharper saliency maps.} of $\pi$. With these quantities, we define our saliency metric for image location $(i,j)$ at time $t$ as
\begin{align}
\mathcal{S}_{\pi}(t,i,j) &= \frac{1}{2} \| \pi_u(I_{1:t}) - \pi_u(I'_{1:t}) \|^2 \\
& \textrm{where} \quad I'_{1:k} = \begin{cases} \Phi(I_k,i,j) \quad & \textrm{if $k=t$} \\ I_k \quad & \textrm{otherwise} \end{cases}
\label{eqn:our-saliency}
\end{align}

The difference $\pi_u(I_{1:t}) - \pi_u(I'_{1:t})$ can be interpreted as a finite differences approximation of the directional gradient $\nabla_{\hat v} \pi_u(I_{1:t})$ where the directional unit vector $\hat v$ denotes the gradient in the direction of $I'_{1:t}$. Our saliency metric is proportional to the squared magnitude of this quantity. This intuition suggests how our perturbation method may improve on gradient-based methods. Whereas the unconstrained gradient need not point in a visually meaningful direction, our directional-gradient approximation is constrained in the direction of a local and meaningful perturbation. We hypothesize that this constraint is what makes our saliency maps more interpretable. 

\textbf{Saliency in practice.} With these definitions, we can construct a saliency map for policy $\pi$ at time $t$ by computing $\mathcal{S}_{\pi}(t,i,j)$ for every pixel in $I_{t}$. In practice, we found that computing a saliency score for $i \mod k$ and $j \mod k$ (in other words, patches of $k=5$ pixels) produced good saliency maps at lower computational cost. For visualization, we upsampled these maps to the full resolution of the Atari input frames and added them to one of the three (RGB) color channels.

We use an identical approach to construct saliency maps for the value estimate $V^{\pi}$. In this case, we defined our saliency metric as the squared difference between the value estimate of the original sequence and that of the perturbed one. That is,
\begin{equation}
\mathcal{S}_{V^{\pi}}(t,i,j) = \frac{1}{2} \| V^{\pi}(I_{1:t}) - V^{\pi}(I'_{1:t}) \|^2.
\end{equation}
This provides a measure of each image region's importance to the valuation of the policy at time $t$. Throughout the paper, we will generally display policy network saliency in blue and value network saliency in red. 

\section{Experiments} \label{sec:experiments}


\subsection{Implementation Details} 

All of our Atari agents have the same recurrent architecture. The input at each time step is a preprocessed version of the current frame. Preprocessing consisted of gray-scaling, down-sampling by a factor of 2, cropping the game space to an $80 \times 80$ square and normalizing the values to $[0,1]$. This input is processed by 4 convolutional layers (each with 32 filters, kernel sizes of 3, strides of 2, and paddings of 1), followed by an LSTM layer with 256 hidden units and a fully-connected layer with $n+1$ units, where $n$ is the dimension of the Atari action space. We applied a softmax activation to the first $n$ neurons to obtain $\pi(I_{1:t})$ and used the last neuron to predict the value, $V^{\pi}(I_{1:t})$.

For our first set of experiments, we trained agents on Pong, Breakout, and SpaceInvaders using the OpenAI Gym API \cite{Brockman2016OpenAIGym,Bellemare2013TheAgents}. We chose these environments because each poses a different set of challenges and deep RL algorithms have historically exceeded human-level performance in them \cite{Mnih2015Human-levelLearning}.

We used the A3C RL algorithm \cite{Mnih2016AsynchronousLearning} with a learning rate of $\alpha=10^{-4}$, a discount factor of $\gamma=0.99$, and computed loss on the policy using Generalized Advantage Estimation with $\lambda=1.0$ \cite{Schulman2016HighEstimation}. Each policy was trained asynchronously for a total of 40 million frames with 20 CPU processes and a shared Adam optimizer 
\cite{Kingma2014Adam:Optimization}.

\subsection{Understanding Strong Policies}

Our first objective was to use saliency videos to explain the strategies learned by strong Atari agents. These agents all exceeded human baselines in their environments by a significant margin. First, we generated saliency videos for three episodes (2000 frames each). Next, we conducted a qualitative investigation of these videos and noted strategies and features that stood out.

\begin{figure}[h!]
\begin{centering}

\subfigure[Pong: kill shot I]{\includegraphics[width=.45\columnwidth]{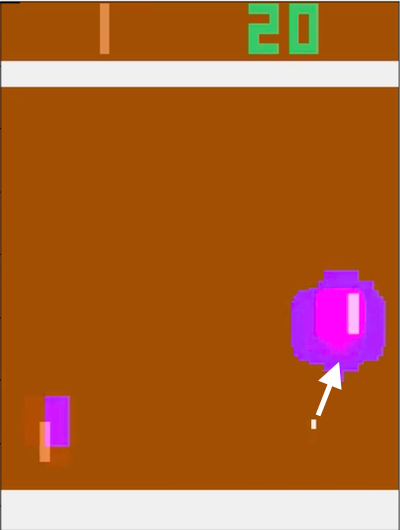}}\hspace{0.15cm}
\subfigure[Pong: kill shot II]{\includegraphics[width=.45\columnwidth]{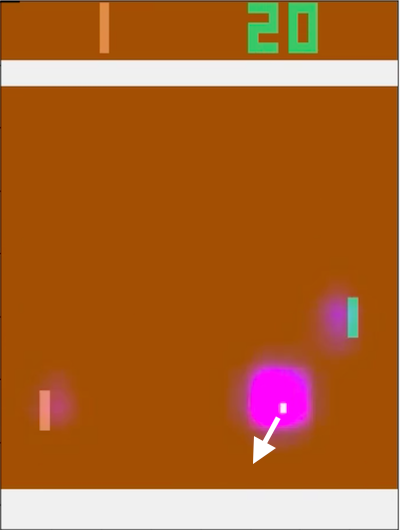}}

\subfigure[SpaceInvaders: aiming I]{\includegraphics[width=.45\columnwidth]{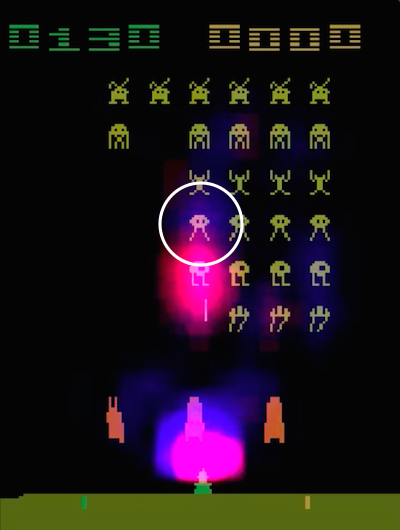}}\hspace{0.15cm}
\subfigure[SpaceInvaders: aiming II]{\includegraphics[width=.45\columnwidth]{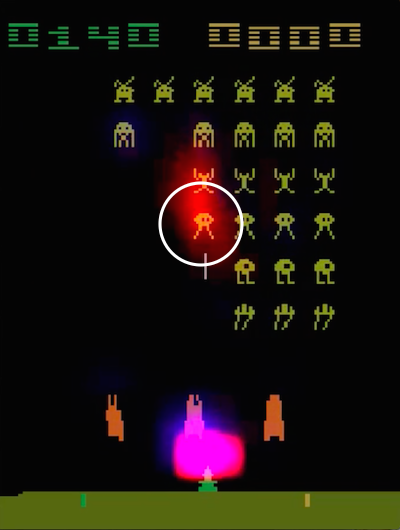}}

\subfigure[Breakout: tunneling I]{\includegraphics[width=.45\columnwidth]{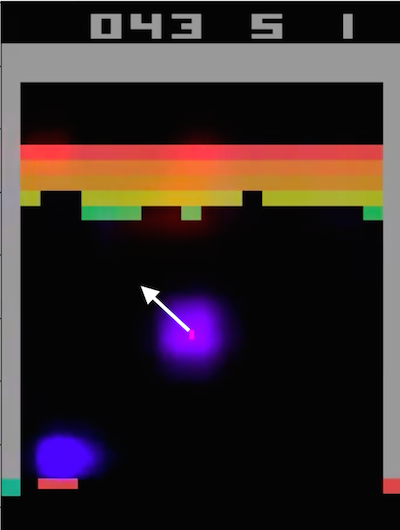}}\hspace{0.15cm}
\subfigure[Breakout: tunneling II]{\includegraphics[width=.45\columnwidth]{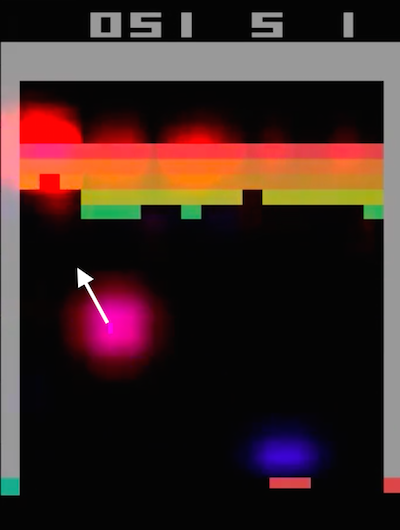}}

\end{centering}
\caption{Visualizing strong Atari 2600 policies. We use an actor-critic network; the actor's saliency map is blue and the critic's saliency map is red. White arrows denote motion of the ball.}
\label{fig:strong}
\end{figure}

\textbf{The strong Pong policy.} Our deep RL Pong agent learned to beat the hard-coded AI over 95\% of the time, often by using a ``kill shot" which the hard-coded AI was unable to return. Our initial understanding of the kill shot, based on observing the policy without saliency, was that the RL agent had learned to first ``lure" the hard-coded AI into the lower region of the frame and then aim the ball towards the top of the frame, where it was difficult to return.

Saliency visualizations told a different story. Figure \ref{fig:strong}a shows a typical situation where the ball is approaching the RL agent's paddle (right side of screen). The agent is positioning its own paddle, which allows it to return the ball at a specific angle. Interestingly, from the saliency we see that the agent attends to very little besides its own paddle: not even the ball. This is because the movements of the ball and opponent are fully deterministic and thus require minimal frame-wise attention. 

After the agent has executed the kill shot (Figure \ref{fig:strong}b), we see that saliency centers entirely around the ball. This makes sense since at this point neither paddle can alter the outcome and their positions are irrelevant. Based on this analysis, it appears that the deep RL agent is exploiting the deterministic nature of the Pong environment. It has learned that it can obtain a reward with high certainty upon executing a precise series of actions. This insight, which cannot be determined by just observing behavior, gives evidence that the agent is not robust and has overfit to the particular opponent. 

\textbf{The strong SpaceInvaders policy.} When we observed our SpaceInvaders agent without saliency maps, we noted that it had learned a strategy that resembled aiming. However, we were not certain of whether it was ``spraying" shots towards dense clusters of enemies or whether it was picking out individual targets.

Applying saliency videos to this agent revealed that it had learned a sophisticated aiming strategy during which first the actor and then the critic would ``track" a target. Aiming begins when the actor highlights a particular alien in blue (circled in Figure \ref{fig:strong}c). This is somewhat difficult to see because the critic network is also attending to a recently-vanquished opponent below. Aiming ends with the agent shooting at the new target. The critic highlights the target in anticipation of an upcoming reward (Figure \ref{fig:strong}d). Notice that both actor and critic tend to monitor the area above the ship. This may be useful for determining whether the ship is protected from enemy fire or has a clear shot at enemies.


\textbf{The strong Breakout policy.} Previous works have noted that strong Breakout agents develop tunneling strategies \cite{Mnih2015Human-levelLearning, Zahavy2016GrayingDQNs}. During tunneling, an agent repeatedly directs the ball at a region of the brick wall in order to tunnel through it. The strategy allows the agent to obtain dense rewards by bouncing the ball between the ceiling and the top of the brick wall. It is unclear how these agents learn and represent tunneling.

A natural expectation is that possible tunneling locations become, and remain, salient from early in the game. Instead, we found that the agent enters and exits a ``tunneling mode" over the course of a single frame. Once the tunneling location becomes salient, it remains so until the tunnel is finished. In Figure \ref{fig:strong}e, the agent has not yet initiated a tunneling strategy and the value network is relatively inactive. Just 20 frames later, the value network starts attending to the far left region of the brick wall, and continues to do so for the next 70 frames (Figure \ref{fig:strong}f).

\subsection{Policies During Learning}

\begin{figure*}[h!]
\begin{centering}
\subfigure[Breakout: learning what features are important.]{\includegraphics[width=1.6\columnwidth]{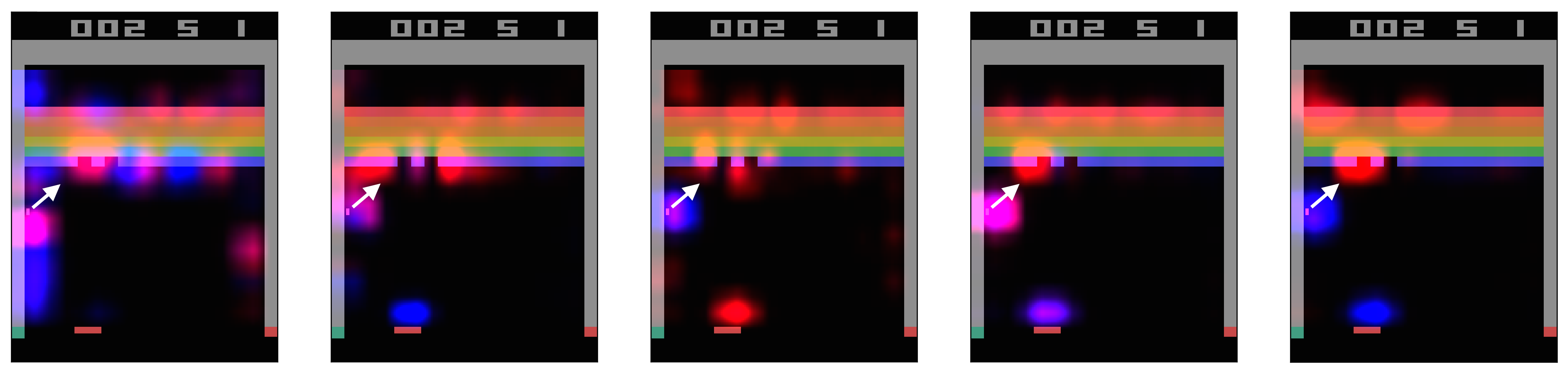}}

\subfigure[Breakout: learning a tunneling strategy.]{\includegraphics[width=1.6\columnwidth]{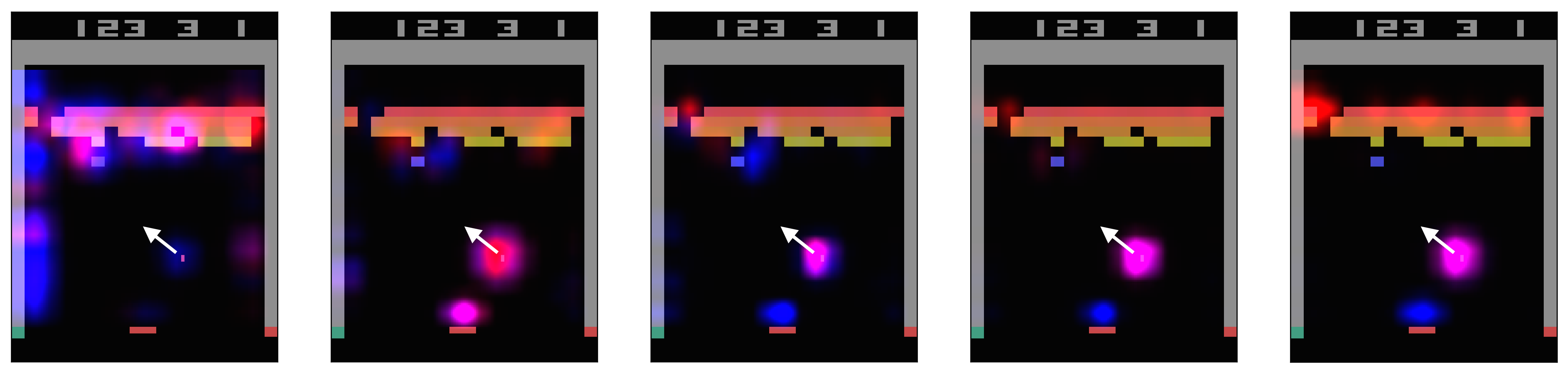}}

\subfigure[Pong: learning a kill shot.]{\includegraphics[width=1.6\columnwidth]{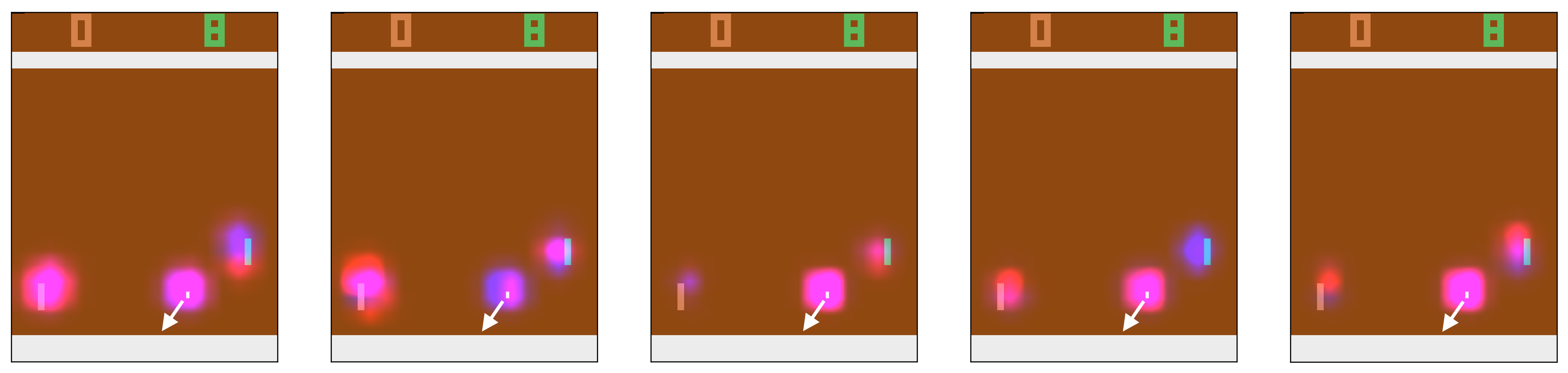}}

\subfigure[SpaceInvaders: learning what features are important and how to aim.]{\includegraphics[width=1.6\columnwidth]{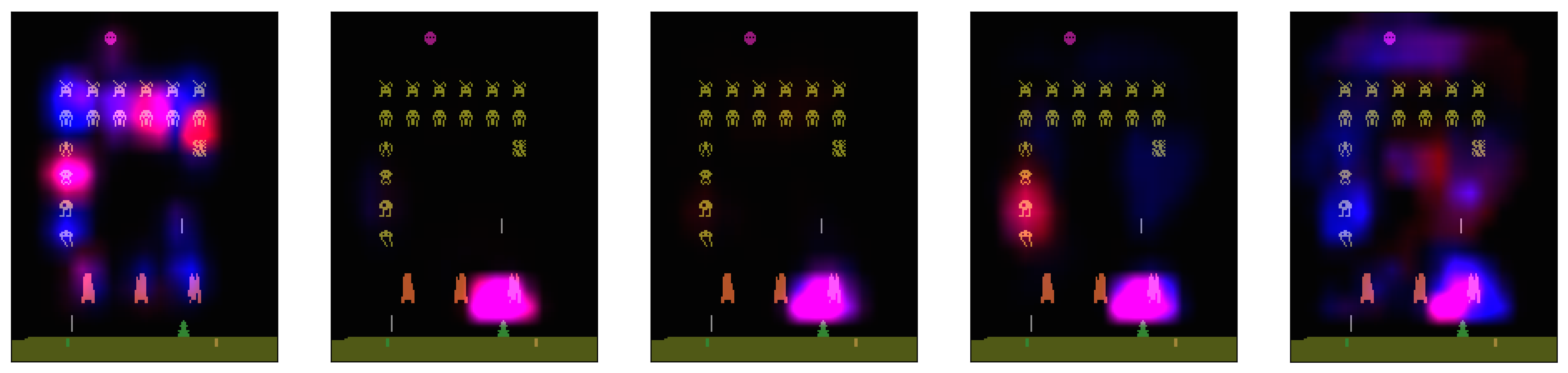}}

\caption{Visualizing learning. Frames are chosen from games played by fully-trained agents. Leftmost agents are untrained, rightmost agents are fully trained. Each column is separated by ten million frames of training. White arrows denote the velocity of the ball.}
\label{fig:learning}
\end{centering}
\end{figure*}

During learning, deep RL agents are known to transition through a broad spectrum of strategies. Some of these strategies are eventually discarded in favor of better ones. Does this process occur in Atari agents? We explored this question by saving several models during training and visualizing them with our saliency method.

\textbf{Learning policies.} Figure \ref{fig:learning} shows how attention changes during the learning process. We see that Atari agents exhibit a significant change in their attention as training progresses. In general, the regions that are most salient to the actor are very different from those of the critic. Figure \ref{fig:learning}b shows that the saliency of the Breakout agent is unfocused during early stages as it learns what is important. As learning progresses, the agent appears to learn about the value of tunneling, as indicated by the critic saliency in the upper left corner. Meanwhile, the policy network learns to attend to the ball and paddle in Figure \ref{fig:learning}a.

In SpaceInvaders, we again see a lack of initial focus. Saliency suggests that the half-trained agents are simply ``spraying bullets" upward without aim. These agents focus on the ``shield" in front of the spaceship, which is relevant to staying alive. As training progressed, the agents shifted to an aiming-based policy, even aiming at the high-value enemy ship at the top of the screen. Pong saliency appears to shift slightly to favor the ball.

\subsection{Detecting Overfit Policies}

Sometimes agents earn high rewards for the wrong reasons. They can do this by exploiting unintended artifacts of their environment and reward functions. We refer to these agents as being ``overfit" to their particular environment and reward function. We were interested in whether our saliency method could help us detect such agents.

\begin{figure}[h!]
\begin{centering}

\subfigure[Pong: control]{\includegraphics[width=.42\columnwidth]{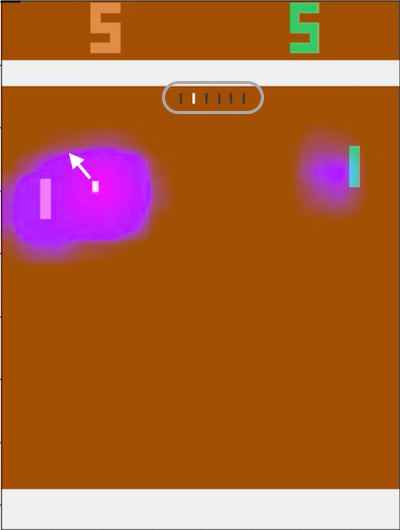}}\hspace{0.15cm}
\subfigure[Pong: overfit]{\includegraphics[width=.42\columnwidth]{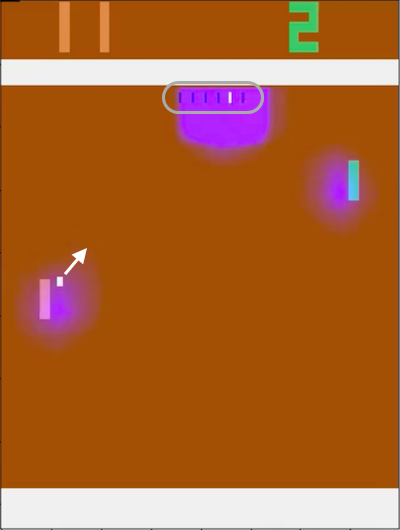}}

\subfigure[SpaceInvaders: control]{\includegraphics[width=.42\columnwidth]{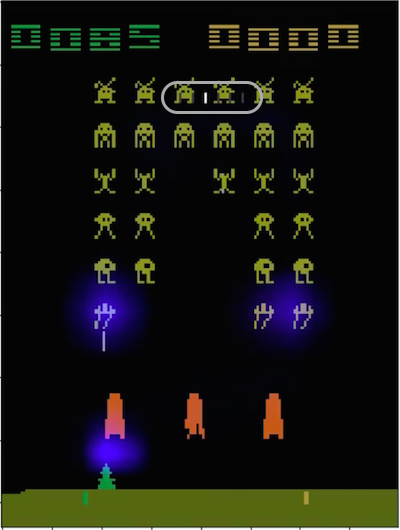}}\hspace{0.15cm}
\subfigure[SpaceInvaders: overfit]{\includegraphics[width=.42\columnwidth]{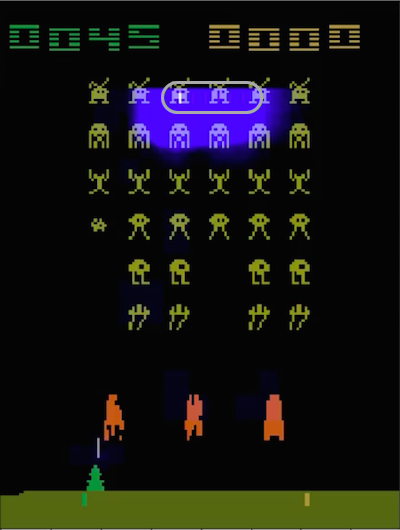}}

\subfigure[Breakout: control]{\includegraphics[width=.42\columnwidth]{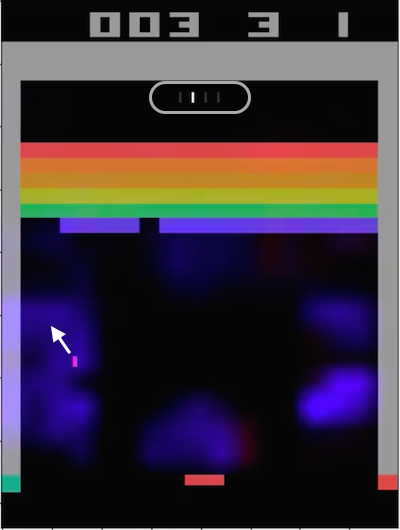}}\hspace{0.15cm}
\subfigure[Breakout: overfit]{\includegraphics[width=.42\columnwidth]{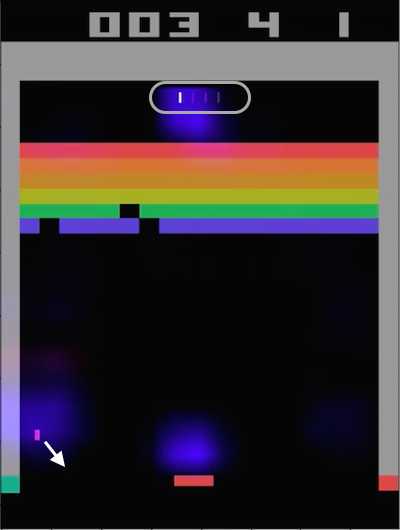}}

\end{centering}
\caption{Visualizing overfit Atari policies. Grey boxes denote the hint pixels. White arrows denote motion of the ball.}
\label{fig:overfit}
\end{figure}

We constructed a toy example where we encouraged overfitting by adding ``hint pixels" to the raw Atari frames. For ``hints" we chose the most probable action selected by a strong ``expert" agent and coded this information as a one-hot distribution of pixel intensities at the top of each frame (see Figure \ref{fig:overfit} for examples).

With these modifications, we trained overfit agents to predict the expert's policy in a supervised manner. We trained ``control" agents in the same manner, assigning random values to their hint pixels. We expected that the overfit agents would learn to focus on the hint pixels, whereas the control agents would need to attend to relevant features of the game space. We halted training after $3\times 10^6$ frames, at which point all agents obtained mean episode rewards at or above human baselines. We were unable to distinguish overfit agents from control agents by observing their behavior alone.

In all three games, our saliency method indicated a clear difference between overfit and control agents. This finding validates our saliency method, in that it can pinpoint regions that we already know to be important. Second, it serves as a good example of how saliency maps can detect agents that obtain high rewards for the wrong reasons.

\subsection{Visualizations for Non-experts}

Convincing human users to trust deep RL agents is a notable hurdle. Non-experts should be able to understand what a strong agent looks like, what an overfit agent looks like, and reason about \textit{why} these agents behave the way they do.

We surveyed 31 students at Oregon State University to measure how our visualization helps non-experts with these tasks. Our survey consisted of two parts. First, participants watched videos of two agents (one control and one overfit) playing Breakout \textit{without} saliency maps. The policies appear nearly identical in these clips. Next, participants watched the same videos \textit{with} saliency maps. After each pair of videos, they were instructed to answer several multiple-choice questions.
\begin{table}[h!]
\centering
\caption{\textit{Which agent has a more robust strategy?}}
\label{tab:robustness}
\begin{tabular}{@{}lccc@{}}
\toprule
                    & Can't tell & Overfit           & Control           \\ \midrule
Video               & 16.1       & \textbf{48.4}     & 35.5              \\
Video + saliency    & 16.1       & 25.8              & \textbf{58.1}     \\ \bottomrule
\end{tabular}
\end{table}

Results in Table \ref{tab:robustness} indicate that saliency maps helped participants judge whether or not the agent was using a robust strategy. In free response, participants generally indicated that they had switched their choice of ``most robust agent" to Agent 2 (control agent) after seeing that Agent 1 (the overfit agent) attended primarily to ``the green dots." 

Another question we asked was \textit{``What piece of visual information do you think Agent X primarily uses to make its decisions?"}. Without the saliency videos, respondents mainly identified the ball (overfit: 67.7\%, control: 41.9\%). With saliency, most respondents said the overfit agent was attending to the hint pixels (67.7\%). Others still chose the ball because, in some frames, the overfit agents attended to both the hint pixels and the ball. The percentage of respondents who identified the ball as the key piece of visual information for the control agent \textit{decreased} to 32.3\% with saliency. This is probably because saliency maps reveal that agents track several objects (the paddle, the ball, and the wall of bricks) simultaneously.

\begin{centering}
\begin{figure*}[h!]
\centering
\subfigure[MsPacman]{\includegraphics[width=.5\columnwidth]{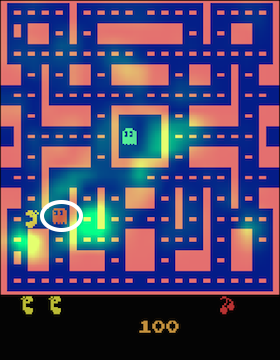}} \hspace{0.2cm}
\subfigure[Frostbite]{\includegraphics[width=.5\columnwidth]{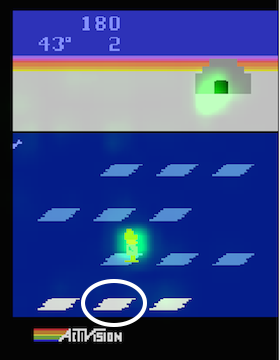}} \hspace{0.2cm}
\subfigure[Enduro]{\includegraphics[width=.5\columnwidth]{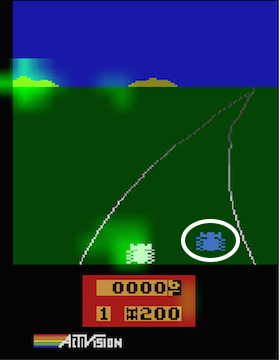}}

\caption{These agents do not attain human performance in the three Atari environments shown. We display the policy saliency in green here because it is easier to see against blue backgrounds. We omit the critic saliency. (a) In MsPacman, the agent should avoid the ghosts. Our agent is not tracking the red ghost, circled. (b) In Frostbite, the agent leaps between platforms. Our agent should attend to its destination platform, circled. Rather, it attends to the goal location at the top of the screen. (c) In Enduro, the agent should avoid other racers. Our agent should be tracking the blue racer, circled. Rather, it focuses on the distant mountains, presumably as a navigation anchor.}
\label{fig:debug}
\end{figure*}
\end{centering}

\subsection{Debugging with Saliency Maps}

In many circumstances, deep RL agents do not converge to good strategies. Even in Atari, there are several environments where our A3C algorithm was never able to surpass human baselines. Examples of these environments include MsPacman, Frostbite, and Enduro (see Figure \ref{fig:debug}). In these cases, saliency maps can help us gain insight into the shortcomings of these policies.

Consider the MsPacman example. As the agent explores the maze, it removes dots from its path, altering the appearance of corridors it has visited. Several of our partially-trained agents appeared to be tracking corridors as a proxy for the PacMan's location. Meanwhile, they \textit{did not} track the ghosts or the PacMan icon (see Figure \ref{fig:debug}). For this reason, the agents were unable to avoid ghosts as they should. This observation led us to examine the reward structure of PacMan; we noticed that the agent was receiving a reward of zero when caught by a ghost. Humans can infer that being caught by a ghost is inherently bad, but the reward structure of MsPacman appears to be too sparse for our agent to make the same inference.

We saw similar patterns in the other two environments, which we explain in Figure \ref{fig:debug}. In both cases, the policies appear to be stuck focusing on distractor objects that prevent the agent from performing well. Without saliency, it would be difficult or impossible to understand the particular flaws in these policies. It is an interesting point of future work to leverage such insights to provide guidance to RL agents.

\subsection{Importance of Memory}
\begin{figure}[h!]
\centering
\includegraphics[width=.9\columnwidth]{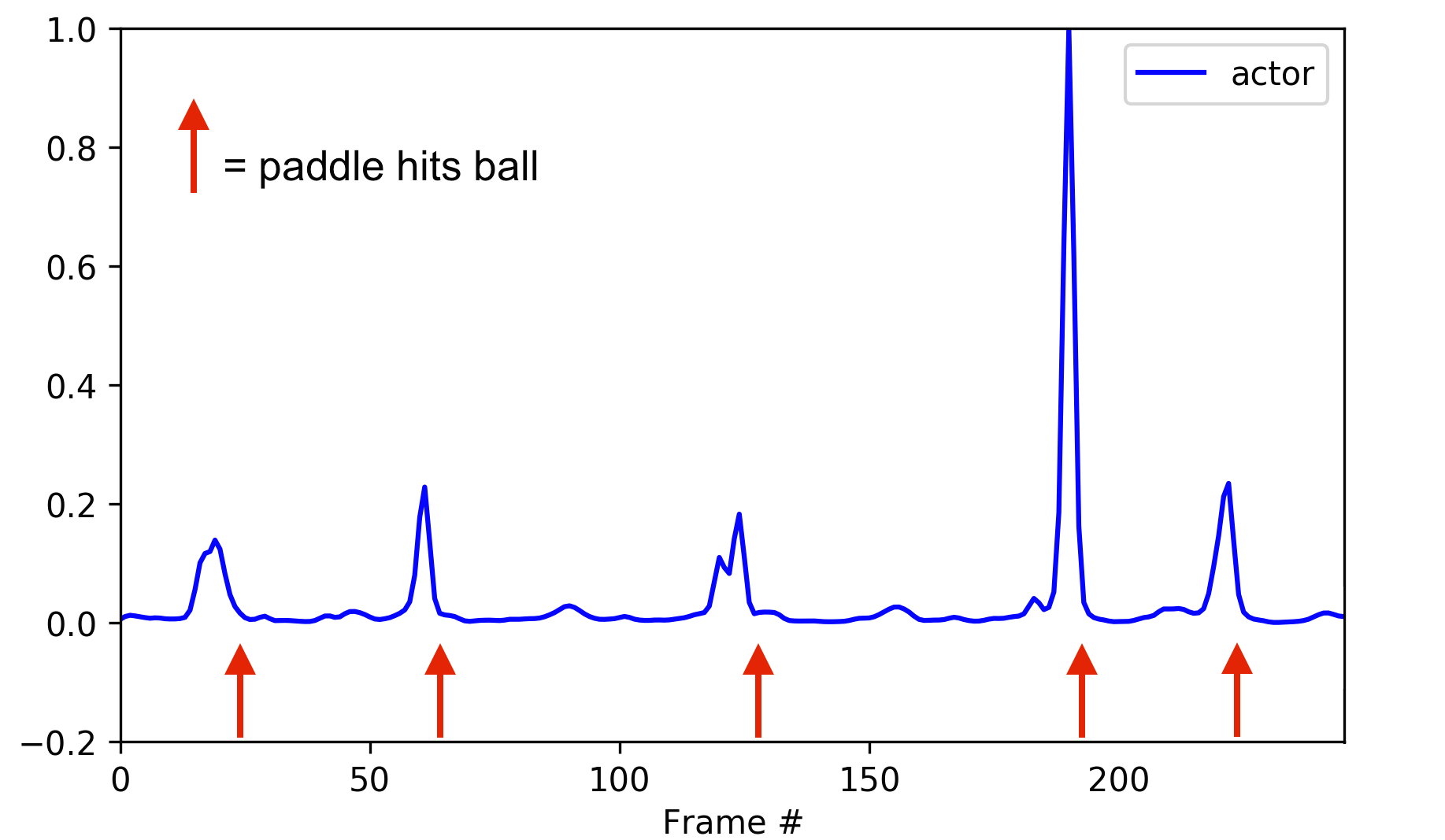}
\caption{Applied to an agent's memory vector, our saliency metric suggests memory is salient just before the ball contacts the paddle.}
\label{fig:breakout-mem}
\end{figure}

Memory is one key part of recurrent policies that we have not addressed. To motivate future directions of research, we modified our perturbation to measure the saliency of memory over time. Since LSTM cell states are not spatially correlated, we chose a different perturbation: decreasing the magnitude of these vectors by 1\%. This reduces the relative magnitude of the LSTM cell state compared to the CNN vector that encodes the input; the degree to which this perturbation alters the policy distribution is a good proxy for the importance of the cell state memory.

Our results suggest that memory is most salient to Pong and Breakout agents immediately before the ball contacts the paddle (see Figure \ref{fig:breakout-mem}). The role of memory in SpaceInvaders was less clear. These results are interesting but preliminary and we recognize that the policy might be most sensitive to \textit{any} perturbations immediately before the paddle contacts the ball. Understanding the role of memory in these agents may require very different types of visualization tools.

\section{Summary}

In this paper, we addressed the growing need for human-interpretable explanations of deep RL agents by introducing a saliency method and using it to visualize and understand Atari agents. We found that our method can yield effective visualizations for a variety of Atari agents. We also found that these visualizations can help non-experts understand what agents are doing. Yet to produce explanations that satisfy human users, researchers will need to use not one, but many techniques to extract the ``how" and ``why" of policies. This work compliments previous efforts, taking the field a step closer to producing truly satisfying explanations.


\section*{Acknowledgments}

This material is based upon work supported by the Defense Advanced Research Projects Agency (DARPA) under Contract N66001-17-2-4030. Any opinions, findings and conclusions or recommendations expressed in this material are those of the authors and do not necessarily reflect the views of the DARPA, the Army Research Office, or the United States government.

Thanks to James and Judy Greydanus for feedback on early versions of this paper.

\bibliography{visualize-atari}
\bibliographystyle{icml2018}
\end{document}